\def\year{2022}\relax
\documentclass[letterpaper]{article} 
\usepackage{aaai22}  
\usepackage{times}  
\usepackage{helvet}  
\usepackage{courier}  
\usepackage[hyphens]{url}  
\usepackage{graphicx} 
\usepackage{natbib}  
\usepackage{caption} 
\DeclareCaptionStyle{ruled}{labelfont=normalfont,labelsep=colon,strut=off} 
\usepackage{booktabs}
\usepackage{multirow}
\usepackage{array}
\usepackage{amssymb}
\usepackage{algorithm}
\usepackage{algorithmic}
\usepackage{bm}
\usepackage{amsmath}
\usepackage{graphicx}

\makeatletter
\DeclareRobustCommand{\cev}[1]{%
  {\mathpalette\do@cev{#1}}%
}
\newcommand{\do@cev}[2]{%
  \vbox{\offinterlineskip
    \sbox\z@{$\m@th#1 x$}%
    \ialign{##\cr
      \hidewidth\reflectbox{$\m@th#1\vec{}\mkern4mu$}\hidewidth\cr
      \noalign{\kern-\ht\z@}
      $\m@th#1#2$\cr
    }%
  }%
}
\makeatother

\usepackage{newfloat}
\usepackage{listings}
\floatstyle{ruled}
\newfloat{listing}{tb}{lst}{}
\floatname{listing}{Listing}
\title{A Label Dependence-aware Sequence Generation Model for \\ Multi-level Implicit Discourse Relation Recognition}
\author {
    Changxing Wu\textsuperscript{\rm 1},
    Liuwen Cao\textsuperscript{\rm 1},
    Yubin Ge\textsuperscript{\rm 2},
    Yang Liu\textsuperscript{\rm 3},
    Min Zhang\textsuperscript{\rm 4},
    Jinsong Su\textsuperscript{\rm 5}\footnote{Corresponding author}
}
\affiliations {
    \textsuperscript{\rm 1} School of Software, East China Jiaotong University, Nanchang, China\\
    \textsuperscript{\rm 2} University of Illinois at Urbana-Champaign, Urbana, IL 61801, USA\\
    \textsuperscript{\rm 3} Tsinghua University, Beijing, China~~~
    \textsuperscript{\rm 4} Soochow University, Soochow, China~~~
    \textsuperscript{\rm 5} Xiamen University, Xiamen, China\\
    wuchangxing@ecjtu.edu.cn, caoliuwencc@163.com, yubinge2@illinois.edu,\\
    liuyang2011@tsinghua.edu.cn, minzhang@suda.edu.cn, jssu@xmu.edu.cn
}

\usepackage{bibentry}

\begin{document}

\maketitle

\begin{abstract}

Implicit discourse relation recognition (IDRR) is a challenging but crucial task in discourse analysis.
Most existing methods train multiple models to predict multi-level labels independently,
while ignoring the dependence between hierarchically structured labels.
In this paper,
we consider multi-level IDRR as a conditional label sequence generation task
and propose a \textbf{L}abel \textbf{D}ependence-aware \textbf{S}equence \textbf{G}eneration \textbf{M}odel (LDSGM) for it.
Specifically,
we first design a label attentive encoder to learn the global representation of an input instance and its level-specific contexts,
where the label dependence is integrated to obtain better label embeddings.
Then,
we employ a label sequence decoder to output the predicted labels in a top-down manner,
where the predicted higher-level labels are directly used to guide the label prediction at the current level.
We further develop a mutual learning enhanced training method to exploit the label dependence in a bottom-up direction,
which is captured by an auxiliary decoder introduced during training.
Experimental results on the PDTB dataset show that our model
achieves the state-of-the-art performance on multi-level IDRR.
We will release our code at https://github.com/nlpersECJTU/LDSGM.

\end{abstract}


\section{Introduction}

\begin{figure}[t]
\centering
\includegraphics[width=8.5cm,height=3.5cm]{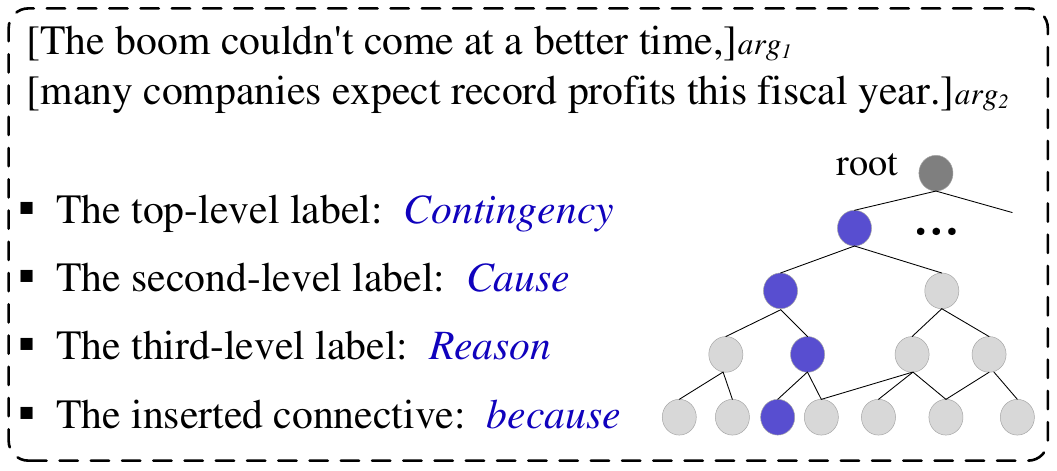}
\caption{An implicit PDTB instance with annotated multi-level labels and the inserted connective.
         Labels defined in PDTB are organized in a three-layer hierarchical structure,
         and the inserted connectives can be considered as the fourth-level (most fine-grained) labels.
         For an instance, its multi-level labels and the inserted connective can be considered as a label sequence,
         e.g., \emph{Contingency}, \emph{Cause}, \emph{Reason}, \emph{because} (the path of blue nodes) for the above instance.}
\label{fig1}
\end{figure}

As a crucial task in discourse analysis,
implicit discourse relation recognition (IDRR) aims to identify semantic relations (e.g., \emph{Contingency})
between two arguments (sentences or clauses).
Figure \ref{fig1} shows an instance from PDTB \cite{prasad_penn_2008}.
It consists of two arguments ($arg_1$ and $arg_2$) and is annotated with three hierarchical labels\footnote{
It is noteworthy that the organization of discourse relation labels is predominantly hierarchical,
which holds in the corpora of different languages, for example, CDTB \cite{li_building_2014}.},
where the second-level/child label \emph{Cause} further refines the semantic of the top-level/parent label \emph{Contingency}, and so on.
During annotation, an implicit connective \emph{because} is first inserted to benefit the label annotation,
and can be considered as the most fine-grained label.
Compared with explicit discourse relation recognition,
IDRR is more challenging because of the absence of explicit connectives.
Due to its wide applications in many natural language processing (NLP) tasks,
such as summarization \cite{cohan_discourse-aware_2018} and event relation extraction \cite{tang_discourse_2021},
IDRR has become one of hot research topics recently.

With the rapid development of deep learning,
previous IDRR models relying on human-designed features have evolved to neural network based models \cite{zhang_shallow_2015,liu_recognizing_2016,lei_swim:_2017,xu_topic_2019,liu_importance_2020,zhang_context_2021}.
Despite their success,
these studies usually train multiple models to predict multi-level labels independently,
while ignoring the dependence between these hierarchically structured labels.
To address this issue,
\citet{nguyen_employing_2019} defined three loss terms to leverage the dependence between labels and connectives.
Though achieving promising results,
they only exploit the label-connective dependence to guide the model training.
\citet{wu_hierarchical_2020} introduced a hierarchical multi-task neural network stacked with a CRF layer to infer the sequence of multi-level labels.
However, there still exist two drawbacks:
1) They stack several feature layers together,
and directly use features from different layers for label predictions at different levels,
which may be insufficient to fully exploit the label dependence for feature extraction.
2) During decoding, the CRF layer has the limitation in processing the label dependence because
it only uses transition matrices to encourage valid label paths and discourage other paths \cite{collobert_natural_2011}.
In other words, the label predictions at higher levels are not directly used to guide the label prediction at the current level.
Therefore it can be said that the potential of the dependence between hierarchically structured labels has not been fully exploited for multi-level IDRR.

In this paper, we consider multi-level IDRR as a conditional label sequence generation task
and propose a \textbf{L}abel \textbf{D}ependence-aware \textbf{S}equence \textbf{G}eneration \textbf{M}odel (LDSGM).
Overall, our model consists of a label attentive encoder and a label sequence decoder,
both of which can effectively leverage the dependence between labels.
In the encoder, based on the hierarchical structure of labels,
we introduce graph convolutional network \cite{kipf_semi-supervised_2017} to learn better label embeddings,
which are used as the query of attention mechanisms to extract level-specific contexts.
Then the decoder sequentially outputs the predicted labels level by level in a top-down manner.
In this way, the easily-predicted higher-level labels can be directly used to
guide the label prediction at the current level.

We further develop a mutual learning enhanced training method to leverage the label dependence.
Concretely,
we introduce an auxiliary decoder that sequentially conducts label predictions in a bottom-up manner.
The intuition behind is that annotators usually first insert suitable connectives
to facilitate the higher-level label annotations \cite{prasad_penn_2008,li_building_2014}.
Our decoder and the auxiliary one are able to
capture the label dependence in two different directions and thus are complement to each other.
Therefore, we transfer the knowledge of the auxiliary decoder to enhance ours via mutual learning \cite{zhang_deep_2018}.
Please note that the auxiliary decoder is not involved during inference.

The main contributions of our work are three-fold:
\begin{itemize}
\item We consider multi-level IDRR as a label sequence generation task.
      To our knowledge, our work is the first attempt to deal with this task in the way of sequence generation.

\item We propose a label dependence-aware sequence generation model.
      Both its encoder and decoder, combined with the mutual learning enhanced training method,
      can fully leverage the label dependence for multi-level IDRR.

\item Experimental results and in-depth analysis show that our model achieves the state-of-the-art performance on the PDTB dataset,
      and more consistent predictions on multi-level labels.
\end{itemize}

\section{Our Model}

\begin{figure*}[t]
\centering
\includegraphics[width=16.0cm,height=11.5cm]{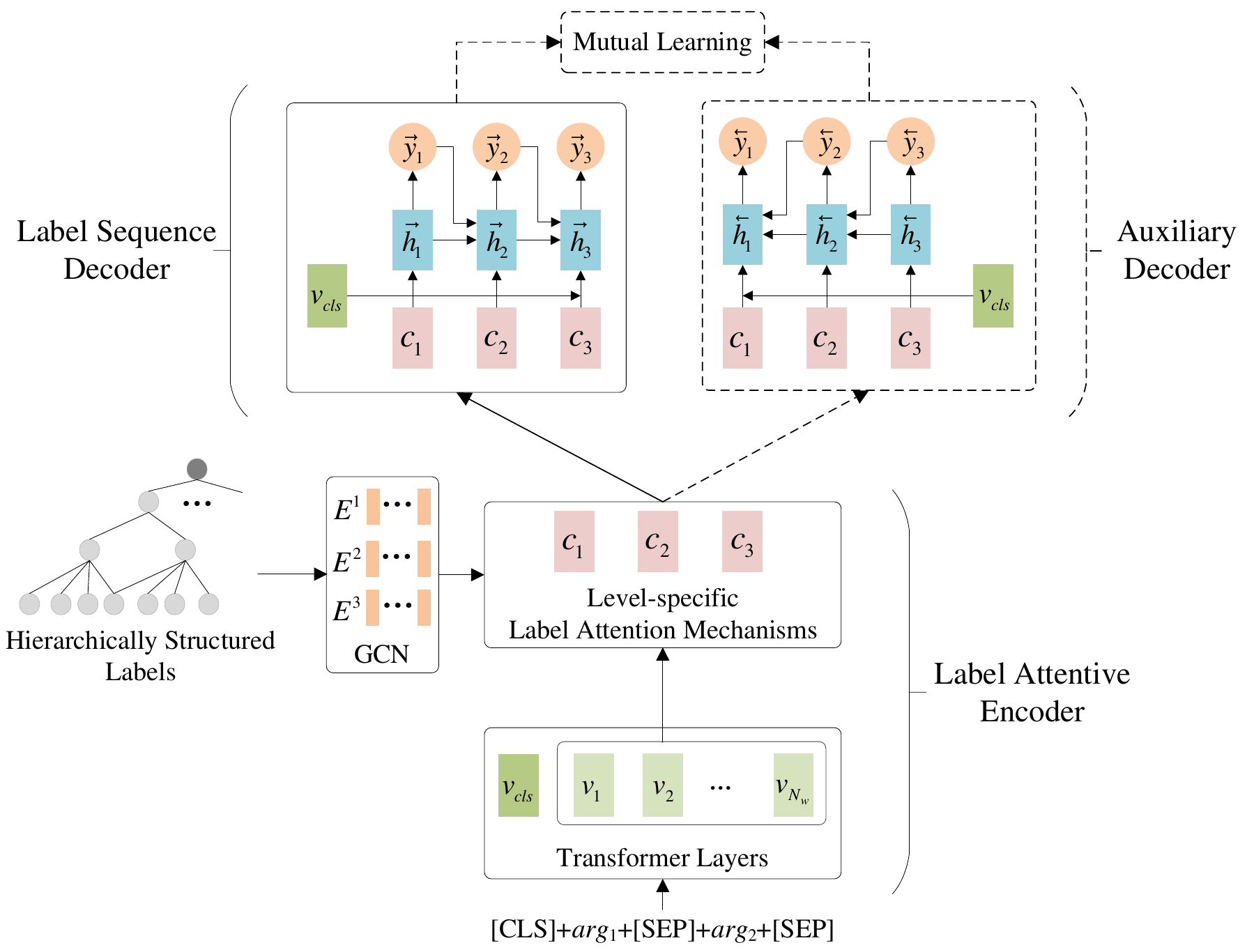}
\caption{The architecture of our LDSGM model.
         The dotted parts are only included at the training phase.
        }
\label{fig2}
\end{figure*}

Given $M$ hierarchical levels of defined labels $\mathbb{C}=(C^{1}, ..., C^m, ..., C^M)$,
where $C^m$ is the set of labels in the $m$-th hierarchical level,
and an instance ${{\bm {{\rm {x}}}}}\!=\!(arg_1, arg_2)$ as input,
our LDSGM model aims to generate a label sequence ${\bm {{\rm {y}}}}={\rm {y}}_1,...,{\rm {y}}_m,...,{\rm {y}}_M$,
where ${\rm{y}}_m\!\in\!C^m$.
As shown in Figure \ref{fig2},
our model mainly consists of a label attentive encoder and a label sequence decoder.
In the following sections,
we will detail these two components
and the mutual learning enhanced training method for our model.

\subsection{Label Attentive Encoder}

Our encoder comprises several stacked Transformer layers,
a graph convolutional network (GCN) and level-specific label attention mechanisms.
Specifically, we employ the Transformer layers to learn the local and global representations of an input instance,
the GCN to obtain better label embeddings by integrating the dependence between hierarchically-structured labels,
and finally the label attention mechanisms to extract level-specific contexts from the local representations.
Afterwards, the learned global representation and level-specific contexts are used as inputs of our decoder for label sequence generation.

Given an instance ${{\bm {{\rm {x}}}}}\!\!=\!\!(arg_1, arg_2)$,
we first stack $K$ Transformer layers \cite{vaswani_attention_2017} to learn its word-level semantic representations as follows:
\begin{equation}
\label{eq:transformer}
\begin{array}{c}
v_{cls}, v_1, ..., v_i, ..., v_{N_{{w}}}\!\!=\!\!{\rm {Transformers}}(arg_1, arg_2),
\end{array}
\end{equation}
where the instance is formulated as a sequence with special tokens: [CLS]+$arg_1$+[SEP]+$arg_2$+[SEP],
$v_{cls}$ is the representation of [CLS],
and ${\{v_i\}}_{i=1}^{N_{{w}}}$ are representations of other tokens.
The key part of a Transformer layer is the multi-head self-attention mechanism.
For IDRR, by concatenating two arguments as input,
this mechanism can directly capture the semantic interactions
between words within not only the same argument but also different arguments.
During the subsequent predictions,
$v_{cls}$ is considered as the global representation of the instance
and thus involved for the label predictions at each level,
while ${\{v_i\}}_{i=1}^{N_{{w}}}$ are local representations from which we extract level-specific contexts for predictions at different levels.

Besides, we adopt a GCN to learn better label embeddings based on the hierarchical structure of labels.
GCN is a multi-layer graph-based neural network
that induces node embeddings based on properties of their neighborhoods \cite{kipf_semi-supervised_2017}.
In this work, we directly consider the hierarchical structure of labels as an undirected graph,
where each label corresponds to a graph node.
The corresponding adjacent matrix $A \in \mathbb R^{N_{C} \times N_{C}} $ is defined as follows:
\begin{equation}
\label{eq:adj}
\begin{array}{c}
{A_{jk}} = \left\{ {\begin{array}{*{20}{c}}
   1 & {if \; {j = k}}    \\
   1 & {if \; {\rm {Parent}}(j)\!=\!k \; {\rm{or}} \; {\rm {Parent}}(k)\!=\!j},\\
   0 & {\rm{otherwise}} \\
\end{array}} \right.
\end{array}
\end{equation}
where ${\rm {Parent}}(j)\!\!=\!\!k$ means that the label corresponding to node $j$ is the subclass of that to node $k$,
and $N_{C}$ is total number of nodes (labels).
Using GCN, given the initial representation of node $j$ as $e_j^{0}$,
we produce the output $e^{l}_j$ of node $j$ at the $l$-th layer ($l \in [1, L]$) as
\begin{equation}
\label{eq:gcn}
\begin{array}{c}
e^{l}_j= \sigma (\sum\limits_{k = 1}^{N_{C}} {{A_{jk}}{W^{l}}e_k^{l-1} + {b^{l}}} ),
\end{array}
\end{equation}
where $W^{l}$ and $b^{l}$ are the parameters,
$\sigma(*)$ is a nonlinear function.
Finally, we obtain the label embeddings as the outputs $\{ {e^{L}_j} \} _{j=1}^{N_C} $ of the $L$-th layer.
For simplicity, we denote all label embeddings at the $m$-th level as $E^m \in \mathbb{R}^{d_e \times |C^m|}$,
with $d_e$ as the dimension of label embedding.

Our GCN is a simple network with only parameters $W^l \in \mathbb{R}^{d_e \times d_e}$  and $b^l \in \mathbb{R}^{d_e}$ at the $l$-th layer.
We select this simple GCN because discourse label graphs are usually small,
for example, only 117 nodes and 262 edges in our experiments.
It is also noteworthy that the label hierarchy is a graph (not tree) structure because many-to-many mappings
exist between the second-level labels and connectives (some connectives are ambiguous).

In our encoder, one of the important steps is to capture the associations between input arguments and labels at each level.
To this end, we design level-specific label attention mechanisms to learn contexts for label predictions.
Specifically, given the label embeddings $E^m$
and the learned local representations $V\!\!=\!\!\{ v_1, ..., v_i, ..., v_{N_{{w}}} \} \in \mathbb R^{d_w \times N_{{w}}}$,
we adopt the attention mechanism at the $m$-th level to extract the level-specific context $c_m$ in the following way:
\begin{equation}
\label{eq:atten}
\renewcommand\arraystretch{1.2}
\begin{aligned}
{c_m} &= V\alpha,  \\
{\alpha} &= {{\rm {softmax}} (\rm {max\text{-}pooling}} (V^{\top}W^m E^m)), \\
\end{aligned}
\end{equation}
where $W^m\!\in\!\mathbb{R}^{d_w \times d_e}$ is a parameter matrix,
$\alpha\!\in\!\mathbb{R}^{N_w}$ is the calculated weight vector,
${\rm {max\text{-}pooling}}(*)$ is the column-wise max-pooling operator,
and $\top$ denotes the matrix transpose operation.
The $i$-th element in $\alpha$ indicates the association degree between the $i$-th local representation and all label embeddings at the $m$-th level.
Note that all label embeddings at a level are used to query its level-specific context.
Finally, we use the learned contexts $\{c_m\}_{m=1}^M$ for the label predictions at corresponding levels, respectively.

Overall, our encoder has several advantages:
1) The dependence between hierarchically structured labels is fully leveraged for better label embeddings.
2) By using the label attention mechanisms,
our encoder extracts label-aware level-specific contexts for the predictions at different levels.

\subsection{Label Sequence Decoder}

Our decoder is an RNN-based one that sequentially generates the predicted labels in a top-down manner,
that is, the top-level label, the second-level label, and so on.
By doing so,
the easily-predicted higher-level labels can be used to guide the label prediction at the current level.
We choose Gate Recurrent Unit (GRU) \cite{cho_learning_2014} to construct our decoder
because of its wide usage in text generation and the short length of label sequences.
Please note that Transformer can also be used as our decoder.

Formally, given three kinds of inputs, including
the global representation $v_{cls}$,
the extracted level-specific context $c_m$,
and the previously-predicted label distribution ${\vec {{y}}_{m-1}}$,
our decoder produces the distribution ${\vec {{y}}_{m}}$ at the $m$-th level as follows:
\begin{equation}
\label{eq:softmax}
\begin{aligned}
{\vec {{y}}_{m}} &= {\rm {softmax}} ({W_o}{\vec h_m} + {b_o}), \\
{\vec h_m} &= {\rm {GRU}}({\vec h_{m - 1}},[v_{cls};{c_m};g({\vec {{y}}_{m-1}})]), \\
\end{aligned}
\end{equation}
where $\vec h_m$ and $\vec h_{m-1}$ are the decoder hidden states,
$[;]$ is the concatenation operation,
and $W_o$ and $b_o$ are the trainable parameters.
As for $g({\vec {{y}}_{m-1}})$,
only using the label with the highest probability under the distribution ${\vec {{y}}_{m-1}}$ may cause error propagation.
To mitigate this issue,
we follow \citet{yang_sgm_2018} to fully exploit information in ${\vec {{y}}_{m-1}}$
by defining $g({\vec {{y}}_{m-1}})$ as $E^{m-1}{\vec {{y}}_{m-1}}$,
where $E^{m-1}$ represents all label embeddings at the level $m{\rm{-}}1$.
Our decoder exploits high-level predictions in a soft constraint manner.
That is to say,
it does not constrain the current predicted label to be a subclass of the previously-predicted label.

The CRF-based IDRR model \cite{wu_hierarchical_2020} first computes label predictions at all levels simultaneously,
and then infers the globally optimized label sequence.
It does not leverage the predicted higher-level labels when computing the label prediction at current level.
By contrast,
our decoder exploits the predicted label distributions at higher levels for the current prediction,
as described in Equation \ref{eq:softmax}.

\subsection{Mutual Learning Enhanced Training}

As described above,
our decoder generates labels in a top-down manner,
which can only exploit the dependence from the predicted higher-level labels,
leaving that from lower levels unexploited.
To address this issue, we develop a mutual learning enhanced training method,
where the label dependence in the bottom-up direction can be utilized to improve the model training.

As shown in the dotted parts of Figure \ref{fig2},
we introduce an auxiliary decoder that possesses the same architecture as the decoder
but sequentially generates predicted label distributions $\cev {{y}}_M, ..., \cev {{y}}_m, ..., \cev {{y}}_1$
in a bottom-up manner\footnote{The auxiliary decoder also uses the learned global representation $v_{cls}$ and the level-specific contexts $\{c_m\}_{m=1}^M$.}.
In this way, the predicted lower-level labels can be used to help the label prediction at the current level.
The basic intuition behind our auxiliary decoder is that annotators usually insert a suitable connective
to help the higher-level label annotations of a given instance.
These inserted connectives can be considered as the most fine-grained labels,
and have been exploited to boost IDRR in previous studies \cite{qin_adversarial_2017,bai_deep_2018,nguyen_employing_2019}.
Apparently, ours and the auxiliary decoder are able to
capture the label dependence in two different directions and thus are complement to each other.
Therefore, during the model training,
we transfer the knowledge of the auxiliary decoder to enhance ours via mutual learning \cite{zhang_deep_2018}.
Please note that our auxiliary decoder is not involved during inference.

\begin{algorithm}
\small
\caption{The training procedure}
\label{alg:algorithm}
\begin{algorithmic}[1]
\STATE \textbf{Input:} Training set $D$, validation set $D'$
\REPEAT
\REPEAT
\STATE Load a batch size of instances $B \in D$
\STATE Generate predicted label distributions $\vec {{y}}_1, ..., \vec {{y}}_m, ..., \vec {{y}}_M$ using the decoder for each instance in $B$
\STATE Generate predicted label distributions $\cev {{y}}_M, ..., \cev {{y}}_m, ..., \cev {{y}}_1$ using the auxiliary decoder for each instance in $B$
\STATE Update $\theta_e, \theta_d$ by minimizing $L(B; \theta_e, \theta_d)$
\STATE Update $\theta_{ad}$ by minimizing $L(B; \theta_{ad})$
\STATE Save the best model according to the average performance at all levels on $D'$
\UNTIL {no more batches}
\UNTIL {convergence}
\end{algorithmic}
\end{algorithm}

To facilitate the understanding of the training procedure, we describe it in Algorithm \ref{alg:algorithm}.
The most notable characteristic is that
ours and the auxiliary decoder can boost each other by iteratively transferring knowledge between them during training.
To do this,
in addition to the conventional cross-entropy based losses,
we introduce two additional losses to minimize the divergence between the predicted label distributions of these two decoders.
Concretely, suppose $\theta_e$, $\theta_d$ and $\theta_{ad}$ to be the parameter sets of the encoder, the decoder, and the auxiliary decoder,
we define the following objective functions over the training data $D$ to update these parameters, respectively:
\begin{equation}
\label{eq:loss}
\begin{split}
L(D; \theta_e, \theta_d) =  & \sum\limits_{({\bm {{\rm {x}}}},{\bm {{\rm {y}}}}) \in D} \sum\limits_{m = 1}^M \{- \mathbb E_{{{{y}}}_{m}} [ \log {\vec {{y}}_{m}} ] \\
                                 &+ \lambda * {{\rm{KL}}({\cev {{y}}_{m}} || {\vec {{y}}_{m}})} \}, \\
L(D; \theta_{ad}) =  & \sum\limits_{({\bm {{\rm {x}}}},{\bm {{\rm {y}}}}) \in D} \sum\limits_{m = 1}^M  \{- \mathbb E_{{{{y}}}_{m}} [ \log {\cev {{y}}_{m}} ] \\
                             &+ \lambda * {{\rm{KL}}({\vec {{y}}_{m}} || {\cev {{y}}_{m}})} \}, \\
\end{split}
\end{equation}
where $y_m$ is the one-hot encoding of the ground-truth label ${\rm{y}}_m\!\in\!{\bm {{\rm {y}}}}$,
${\vec {{y}}_{m}}$ and ${\cev {{y}}_{m}}$ are the predicted label distributions at the $m$-th level by ours and the auxiliary decoder respectively,
$\mathbb E_{{{{y}}_m}}[*]$ is the expected value with respect to ${{y}}_m$,
${\rm{KL}}(*||*)$ is the Kullback-Leibler(KL) divergence function,
and $\lambda$ is the coefficient used to control the impacts of different loss items.
We repeat the above knowledge transfer process, until both loss functions converge.
By doing so,
the label dependence in both top-down and bottom-up directions can be captured to benefit the decoder during the model training.

\section{Experiments}

To investigate the effectiveness of our model,
we conduct experiments on the PDTB 2.0 corpus \cite{prasad_penn_2008}.

\begin{figure*}
\centering
\includegraphics[width=16.0cm,height=4.5cm]{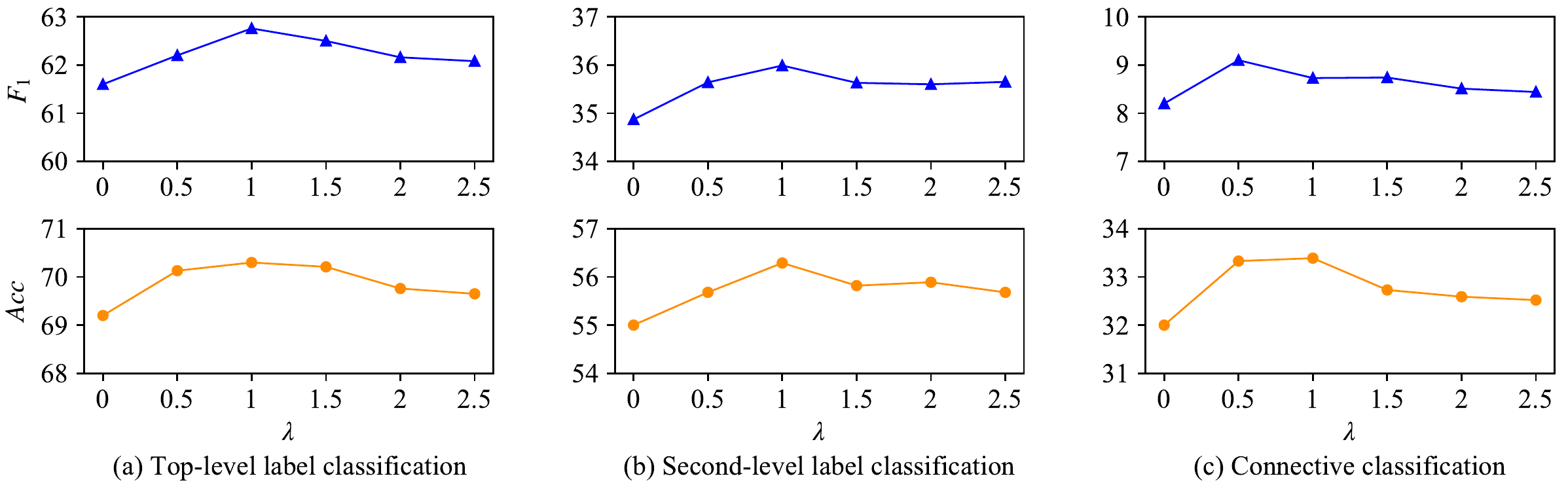}
\caption{Effects of $\lambda$ on the validation set.}
\label{fig:lamb}
\end{figure*}

\subsection{Dataset}

Since only part of PDTB instances are annotated with the third-level labels,
we take the top-level and second-level labels into consideration,
and regard the inserted connectives as the third-level labels.
There are four top-level labels are considered,
including \emph{Temporal (Temp)}, \emph{Contingency (Cont)}, \emph{Comparison (Comp)} and \emph{Expansion (Expa)}.
Further, there exist 16 second-level labels,
five of which with few training instances and no validation and test instance are removed.
Therefore, we conduct an 11-way classification on the second-level labels.
For the connective classification,
we consider all 102 connectives defined in PDTB.
Following previous studies \cite{bai_deep_2018,wu_hierarchical_2020,liu_importance_2020},
we split the PDTB corpus into three parts:
the training set with 12,775 instances (Section 2-20),
the validation set with 1,183 instances (Section 0-1),
and the test set with 1,046 instances (Section 21-22).

\subsection{Settings}

All hyper-parameters are determined according to the best average model performance at three levels on the validation set.
We use the pre-trained RoBERTa-base\footnote{https://huggingface.co/transformers/model\_doc/roberta.html} \cite{liu_roberta:_2019},
which has 12 layers with 768 hidden units and 12 attention heads per layer,
as our Transformer layers.
We set the layer number of GCN as 2 since more layers do not bring the performance improvement.
The graph node embeddings with the dimension 100 are randomly initialized according to the uniform distribution $U[-0.01, 0.01]$.
The GRU hidden states in decoders are 768 dimensions.
All other parameters are initialized with the default values in PyTorch.
To avoid overfitting,
we employ the \emph{dropout} strategy \cite{srivastava_dropout_2014} with the rate 0.2 over the outputs of RoBERTa-base.
Finally, we adopt Adam \cite{kingma_adam_2015} with the learning rate of $e\text{-}5$ and
the batch size of 32 to update the model parameters for 15 epochs.
To alleviate instability in the model training,
we conduct all experiments for five times with different random seeds and report average results.
Due to the skewed distribution of test dataset,
we use both the macro-averaged $F_1$ score and accuracy ($Acc$) to evaluate the model.

\subsection{Baselines}

We compare our LDSGM model with the following competitive baselines:

\begin{itemize}
\item IDRR-C\&E \cite{dai_regularization_2019}: a neural IDRR model using a regularization approach
                                                to leverage coreference relations and external event knowledge.

\item IDRR-Con \cite{shi_learning_2019}: a neural IDRR model attempting to learn better argument representations
                                by leveraging the inserted connectives.

\item KANN \cite{guo_working_2020}: a knowledge-enhanced attentive neural network that effectively fuses discourse context
                               and its external knowledge from WordNet.

\item MTL-MLoss \cite{nguyen_employing_2019}: it is a neural model based on multi-task learning,
                                              where the labels and connectives are simultaneously predicted,
                                    with mapping losses to leverage the dependence between them.

\item BERT-FT \cite{kishimoto_adapting_2020}: it fine-tunes BERT \cite{devlin_bert_2019} for IDRR with texts tailored to discourse classification,
                                      explicit connective prediction and implicit connective prediction.

\item HierMTN-CRF \cite{wu_hierarchical_2020}: a hierarchical multi-task neural network stacked with a CRF layer to infer the sequence of multi-level labels.

\item BMGF-RoBERTa \cite{liu_importance_2020}: a RoBERTa-based model, which is equipped with a bilateral matching module to capture the interaction
                                               between arguments, and a gated fusion module to derive final representations for label classification.

\item MTL-MLoss-RoBERTa and HierMTN-CRF-RoBERTa: for a fair comparison,
                                                we enhance MTL-MLoss and HierMTN-CRF with RoBERTa instead of ELMo and BERT, respectively.

\item OurEncoder+OurDecoder: a variant of our model where the mutual learning enhanced training is removed.

\end{itemize}

\begin{table*}[htp]
\centering
\begin{tabular}{p{7.8cm}<{\centering}|p{1.6cm}<{\centering}|p{0.84cm}<{\centering}|p{0.84cm}<{\centering}|p{0.84cm}<{\centering}|p{0.84cm}<{\centering}|p{0.84cm}<{\centering}|p{0.84cm}<{\centering}}
\toprule
\multirow{2}{*}{Model} &\multirow{2}{*}{Embeddings} &\multicolumn{2}{c|}{Top-level} &\multicolumn{2}{c|}{Second-level} &\multicolumn{2}{c}{Connective}   \\
\cline{3-8}        &             &$F_1$            &$Acc$      & $F_1$             &$Acc$               & $F_1$        &$Acc$     \\
\hline
IDRR-C\&E \cite{dai_regularization_2019}         &GloVe     &50.49            &58.32      &32.13          &46.03              &-             &-        \\
IDRR-Con \cite{shi_learning_2019}          &word2vec     &46.40            &61.42      &-              &47.83              &-             &-        \\
KANN \cite{guo_working_2020}              &GloVe     &47.90            &57.25      &-              &-                  &-             &-        \\
\hline
IDRR-C\&E \cite{dai_regularization_2019}         &ELMo     &52.89            &59.66      &33.41          &48.23              &-             &-        \\
MTL-MLoss \cite{nguyen_employing_2019}         &ELMo     &53.00            &-          &-              &49.95              &-	            &-     \\
BERT-FT \cite{kishimoto_adapting_2020}          &BERT     &58.48            &65.26      &-              &54.32              &-	            &-     \\
HierMTN-CRF \cite{wu_hierarchical_2020}       &BERT     &55.72            &65.26      &33.91      &52.34      &10.37         &30.00     \\
BMGF-RoBERTa \cite{liu_importance_2020}      &RoBERTa   &\underline{63.39}    &69.06      &-        &58.13       &-             &-     \\
MTL-MLoss-RoBERTa   &RoBERTa         &61.89 &68.42 &38.10 &57.72 &7.75 &29.57        \\
HierMTN-CRF-RoBERTa   &RoBERTa       &62.02 &\underline{70.05} &\underline{38.28} &\underline{58.61} &\underline{10.45} &\underline{31.30}         \\
\hline
OurEncoder+OurDecoder      &RoBERTa           &62.93      &70.66       &39.71         &59.59          &10.67          &31.54        \\
LDSGM               &RoBERTa    &\textbf{63.73}   &\textbf{71.18}    &\textbf{40.49}    &\textbf{60.33}    &\textbf{10.68}      &\textbf{32.20}        \\
\bottomrule
\end{tabular}
\caption{Experimental results on PDTB. The best results of previous baselines are underlined.}
\label{table:main}
\end{table*}

\subsection{Effect of the Coefficient $\lambda$}

As shown in Equation \ref{eq:loss},
the coefficient $\lambda$ is an important hyper-parameter
that controls the relative impacts of two different losses.
Thus, we vary $\lambda$ from 0 to 2.5 with an increment of 0.5 each step,
and inspect the performance of our model using different $\lambda$s on the validation set.
The larger $\lambda$ indicates that ours and the auxiliary decoder learn more knowledge from each other.

Figure \ref{fig:lamb} shows the experimental results.
We can find that, compared with the model without mutual learning ($\lambda=0$),
the performance of our model at any level is always improved via mutual learning.
When $\lambda$ exceeding 1.0, the performance of our model tends to be stable and declines finally.
Only one exception is $F_1$ of connective classification, which reaches its best score when $\lambda=0.5$.
Thus, we directly set $\lambda=1.0$ for all experiments thereafter.

\subsection{Main Results}

\begin{table}[htp]
\centering
\begin{tabular}{p{3.7cm}<{\centering}|p{1.4cm}<{\centering}|p{1.9cm}<{\centering}}
\toprule
       Model                         & Top-Sec              & Top-Sec-Con     \\
\hline  HierMTN-CRF                  & 46.29                & 19.15    \\
        BMGF-RoBERTa                 & 47.06                & 21.37   \\
        OurEncoder+OurDecoder        & 57.86                & 25.31   \\
        LDSGM                        & \textbf{58.61}       & \textbf{26.85}   \\
\bottomrule
\end{tabular}
\caption{Comparison with recent models on the consistency among multi-level label predictions.
We run the code of BMGF-RoBERTa \cite{liu_importance_2020} and report the results.}
\label{table:consistent}
\end{table}

Table \ref{table:main} shows the main results,
from which we can reach the following conclusions.
\textbf{First},
all models enhanced with contextualized word embeddings (Parts 2 and 3) outperform those using static word embeddings (Part 1).
\textbf{Second},
in most cases, jointly inferring multi-level labels (HierMTN-CRF-RoBERTa, OurEncoder+OurDecoder, LDSGM)
performs better than separately predicting in BMGF-RoBERTa,
which means that integrating the label dependence is indeed helpful.
\textbf{Third},
our LDSGM model achieves the state-of-the-art performance on classification tasks at all three levels.
When considering accuracy,
it obtains 1.13\%, 1.72\% and 0.90\% improvements over the best results of previous baselines (Part 3), respectively.
In terms of $F_1$, it also performs consistently better than previous models.


\textbf{Consistency among Multi-level Label Predictions}
Unlike most previous studies predicting labels at different levels independently,
our LDSGM model converts multi-level IDRR into a label sequence generation task,
which intuitively is able to alleviate the inconsistency among label predictions at different levels.
To verify this,
we report the model performance via two new metrics:
1) \textbf{Top-Sec}: the percentage of correct predictions at both the top-level and second-level labels;
2) \textbf{Top-Sec-Con}: the percentage of correct predictions across all three levels.
As shown in Table \ref{table:consistent}, compared with baselines,
our model achieves more consistent predictions in terms of both Top-Sec and Top-Sec-Con.

Overall, all above-mentioned results strongly demonstrate the effectiveness of our model in capturing the dependence between hierarchically structured labels.

\subsection{Ablation Study and Analysis}

\begin{table*}[htp]
\centering
\begin{tabular}{p{7.3cm}<{\centering}|p{1.2cm}<{\centering}|p{1.2cm}<{\centering}|p{1.2cm}<{\centering}|p{1.2cm}<{\centering}|p{1.2cm}<{\centering}|p{1.2cm}<{\centering}}
\toprule
\multirow{2}{*}{Model}           &\multicolumn{2}{c|}{Top-level}      &\multicolumn{2}{c|}{Second-level}          &\multicolumn{2}{c}{Connective}     \\
\cline{2-7}                     &$F_1$            &$Acc$      & $F_1$             &$Acc$               & $F_1$        &$Acc$     \\
\hline
LDSGM             &\textbf{63.73}   &\textbf{71.18}    &\textbf{40.49}    &\textbf{60.33}    &\textbf{10.68}      &\textbf{32.20}        \\
~~~~~w/o GCN               &63.47      &70.77       &39.32         &59.54          &9.81          &30.94        \\
~~w/o LA                   &62.84      &70.19       &39.15         &59.09          &10.64          &31.32        \\
~~w/o PP                   &63.68      &70.78       &39.27         &59.62          &10.08          &30.86        \\
w/o ML (aka. OurEncoder+OurDecoder)                   &62.93      &70.66       &39.71         &59.59          &10.67          &31.54        \\
\hline
OurEncoder+Softmax+MultiTask     &62.82    &70.25             &38.28             &58.37             &9.60                &30.82        \\
OurEncoder+CRF          &62.82    &70.53             &38.82             &59.09             &10.41               &31.24        \\
OurEncoder+Ensemble(OurDecoder, AuxDecoder)    &63.04      &70.80       &39.91         &59.61          &8.61           &29.95        \\
\bottomrule
\end{tabular}
\caption{Ablation study.
LA means the level-specific label attention mechanism, PP represents previous predictions, and ML means the mutual learning enhanced training.
Softmax+MultiTask denotes that we stack ${\rm{softmax}}$ layers on the top of our encoder to individually predict labels at different levels,
and then train them based on multi-task learning.
Ensemble(OurDecoder, AuxDecoder) means that we directly ensemble the prediction results of ours and the auxiliary decoder.
}
\label{table:ablation}
\end{table*}

To evaluate the effects of different components,
we compare LDSGM with its variants:
1) \textbf{w/o GCN}. In this variant, the GCN is not used to learn better label embeddings,
which means that the label dependence is not exploited by our encoder.
2) \textbf{w/o LA}. We replace the level-specific label attention with the one without label embeddings,
where the hidden states of decoder are used as query to retrieve contexts for label predictions;
3) \textbf{w/o PP}. In this variant,
the previous predictions are not used to guide the prediction of labels at the current level,
that is to say, $g({\vec {{y}}_{m-1}})$ is removed from Equation \ref{eq:softmax};
4) \textbf{w/o ML}.
Mutual learning enhanced training is not involved,
meaning that our model degenerates into the OurEncoder+OurDecoder model.

From Part 1 of Table \ref{table:ablation},
we can observe that our LDSGM model always exhibits better performance than their corresponding variants across all three levels.
Therefore, we confirm that incorporating label dependence into both our encoder and decoder is indeed beneficial.
Moreover, as shown in Part 2,
with the same encoder,
our LDSGM model always achieves better performance than those based on other decoding schemes,
including Softmax+MultiTask, CRF, and Ensemble(OurDecoder, AuxDecoder).
These results strongly demonstrate the effectiveness of exploiting the label dependence in two different directions (top-down and bottom-up).
Finally,
we should note that the ensemble scheme yields poor performance on connective classification.
The reason behind is that, for the auxiliary decoder,
it is very hard to predict connectives without the guidance of predicted higher-level labels,
with only 6.79\% $F_1$ score and 27.90\% $Acc$.

\begin{table}[htp]
\centering
\begin{tabular}{p{3.2cm}<{\centering}|p{2.6cm}<{\centering}|p{1.2cm}<{\centering}}
\toprule
        Second-level Label                 & BMGF-RoBERTa  & LDSGM         \\
\hline  \emph{Temp.Asynchronous}           & 56.18                       & \textbf{56.47}           \\
        \emph{Temp.Synchrony}              & 0.0                         & 0.0    \\
\hline  \emph{Cont.Cause}                  & 59.60                       & \textbf{64.36}           \\
        \emph{Cont.Pragmatic cause}        & 0.0                         & 0.0    \\
\hline  \emph{Comp.Contrast}               & 59.75                       & \textbf{63.52}           \\
        \emph{Comp.Concession}             & 0.0                         & 0.0    \\
\hline  \emph{Expa.Conjunction}            & 60.17                       & 57.91           \\
        \emph{Expa.Instantiation}          & 67.96                       & \textbf{72.60}           \\
        \emph{Expa.Restatement}            & 53.83                       & \textbf{58.06}           \\
        \emph{Expa.Alternative}            & 60.00                       & \textbf{63.46}           \\
        \emph{Expa.List}                   & 0.0                         & \textbf{8.98}           \\
\bottomrule
\end{tabular}
\caption{Label-wise $F_1$ scores for the second-level labels.}
\label{table:second}
\end{table}

\subsection{Performance on Minority Label Predictions}
\label{sec:minority}

Note that recent studies usually ignore the problem of addressing the highly skewed distribution of labels.
For example,
only about 6.8\% of PDTB instances are annotated as five of the considered 11 second-level labels
(\emph{Temp.Synchrony}, \emph{Cont.Pragmatic cause}, \emph{Comp.Concession}, \emph{Expa.Alternative}, and \emph{Expa.List}).
Therefore, we report the label-wise $F_1$ scores for the second-level labels in Table \ref{table:second}.
A closer look into the results reveals that
though our LDSGM model outperforms BMGF-RoBERTa \cite{liu_importance_2020} on most majority labels,
the $F_1$ scores for three minority labels are still 0\%.
Besides, the BERT-based model \cite{kishimoto_adapting_2020} also shows poor performance on minority labels.
This is because that small numbers of training examples are insufficient
to optimize the huge amounts of parameters in these models.
In the future, we will pay more attention to the minority labels.

\section{Related Work}

Early studies for IDRR mainly resort to manually-designed features \cite{lin_recognizing_2009,park_improving_2012,rutherford_discovering_2014}.
With the rapid development of deep learning,
neural network based IDRR models have gradually become dominant.
In this aspect,
many efforts have been devoted to learning better semantic representations of arguments,
where typical studies include the shallow CNN \cite{zhang_shallow_2015},
entity-enhanced recursive neural network \cite{ji_one_2015},
CNN with character-enhance embeddings \cite{qin_implicit_2016},
variational generation model \cite{zhang_variational_2016},
knowledge-augmented LSTM \cite{kishimoto_knowledge-augmented_2018},
and TransS-driven joint learning framework \cite{he_transs-driven_2020}.
Recently,
more researchers focus on developing neural networks to better capture the semantic interaction between two input arguments for IDRR,
for example, multi-level attention mechanisms \cite{liu_recognizing_2016},
bi-attention mechanisms \cite{lei_swim:_2017},
topic tensor networks \cite{xu_topic_2019},
the knowledge-based memory network \cite{guo_working_2020},
and the graph-based context model \cite{zhang_context_2021}.
To enrich the limited training data,
another line of research focuses on leveraging explicit discourse data
via data selection methods \cite{wu_bilingually-constrained_2016,xu_using_2018},
multi-task learning \cite{liu_implicit_2016,lan_multi-task_2017},
task-specific embeddings \cite{braud_learning_2016,wu_improving_2017},
or pre-training methods \cite{kishimoto_adapting_2020}.
More recently, contextualized representations learned from pre-trained  models,
such as ELMo \cite{peters_deep_2018}, BERT, and RoBERTa,
have significantly improved the performance of IDRR \cite{bai_deep_2018,shi_next_2019,liu_importance_2020}.
All the above studies predict labels at different levels independently.
Unlike them, our work infers multi-level labels jointly via a sequence generation model.
The most related to ours is \cite{wu_hierarchical_2020},
which adopts a CRF-based decoder to infer label sequences.
By contrast,
our work is the first attempt to recast multi-level IDRR as a conditional label sequence generation task.

Our work is also related to recent efforts on multi-label text classification,
where the dependence between labels are used to improve the encoder for better feature extraction \cite{xie_neural_2018,huang_hierarchical_2019,zhou_hierarchy-aware_2020},
or the decoder for joint label predictions \cite{yang_sgm_2018,wu_learning_2019}.
Ours significantly differs from the above work in following aspects:
1) Both the encoder and decoder in our model can effectively leverage the label dependence;
2) A bottom-up auxiliary decoder is introduced to boost our top-down decoder via mutual learning.

\section{Conclusion}

In this paper,
we have presented a label dependence-aware sequence generation model,
which can fully leverage the dependence between hierarchically structured labels for multi-level IDRR.
Experimental results and in-depth analyses on the PDTB corpus clearly demonstrate the effectiveness of our model.
To our knowledge, our work is the first attempt to deal with this task in the way of sequence generation.
An interesting future work is to combine advantages of ours and auxiliary decoder
via asynchronous bidirectional decoding \cite{zhang_asynchronous_2018,su_exploiting_2019} during inference.


\section{Acknowledgments}

We would like to thank all the reviewers for their constructive
and helpful suggestions on this paper.
This work was supported by the National Natural Science Foundation
of China (Nos. 61866012, 62172160, 62166018 and 62062034),
the Natural Science Foundation of Fujian Province of China (No. 2020J06001),
the Youth Innovation Fund of Xiamen (No. 3502Z20206059),
and the Natural Science Foundation of Jiangxi Province of China (Nos. 20212ACB212002, 20202BABL202043).

\bibliography{widrr}

\end{document}